\documentclass{article}

\usepackage[preprint,nonatbib]{neurips_2021}

\usepackage{graphicx}
\usepackage{subcaption}
\usepackage{dblfloatfix}
\usepackage{xcolor}
\usepackage{cite}
\usepackage{amssymb}
\usepackage{wrapfig}
\usepackage{enumitem}
\usepackage{csquotes}

\usepackage[utf8]{inputenc} % allow utf-8 input
\usepackage[T1]{fontenc}    % use 8-bit T1 fonts
\usepackage{hyperref}       % hyperlinks
\usepackage{url}            % simple URL typesetting
\usepackage{booktabs}       % professional-quality tables
\usepackage{amsfonts}       % blackboard math symbols
\usepackage{nicefrac}       % compact symbols for 1/2, etc.
\usepackage{microtype}      % microtypography
\usepackage{xcolor}         % colors

\title{Turath-150K: Image Database of Arab Heritage}

\author{%
  Dani Kiyasseh\\
  Department of Computing and Mathematical Sciences\\
  California Institute of Technology\\
  Pasadena, CA, USA\\
  \texttt{dkiyass1@caltech.edu} \\
   \And
  Rasheed El-Bouri\\
  Department of Engineering Science\\
  University of Oxford\\
  Oxford, UK\\
  \texttt{rasheed.el-bouri@eng.ox.ac.uk} \\
}

\begin{document}

\maketitle

\begin{abstract}

Large-scale image databases remain largely biased towards objects and activities encountered in a select few cultures. This absence of culturally-diverse images, which we refer to as the \enquote{hidden tail}, limits the applicability of pre-trained neural networks and inadvertently excludes researchers from under-represented regions. To begin remedying this issue, we curate Turath-150K, a database of images of the Arab world that reflect objects, activities, and scenarios commonly found there. In the process, we introduce three benchmark databases, Turath Standard, Art, and UNESCO, specialised subsets of the Turath dataset. After demonstrating the limitations of existing networks pre-trained on ImageNet when deployed on such benchmarks, we train and evaluate several networks on the task of image classification. As a consequence of Turath, we hope to engage machine learning researchers in under-represented regions, and to inspire the release of additional culture-focused databases. The database can be accessed here: \url{danikiyasseh.github.io/Turath}.

\end{abstract}

\section{Introduction}
% computer vision and image databases
Deep neural networks have exhibited great success in performing various computer vision tasks, such as image classification \cite{Iandola2016}, object detection \cite{Ren2015}, and segmentation \cite{Chen2017}. One of the key factors and driving forces behind the success of such networks is access to large-scale, annotated datasets that consist of samples that are mostly representative of the underlying data distribution. To that end, publicly-available datasets, such as ImageNet~\cite{deng2009imagenet}, SUN~\cite{xiao2010sun}, and Places~\cite{zhou2017places}, attempt to capture a diverse set of images that are reflective of objects and scenarios encountered \enquote{in the wild}. Such images typically belong to categories guided by the WordNet hierarchy \cite{Fellbaum2010} and which are diversified by incorporating various adjectives into search queries (e.g., night, foggy, etc.)

% Gap/ challenge of existing databases
Despite these efforts, existing databases remain largely biased towards objects, activities, and scenarios commonly encountered in a small subset of cultures \cite{Birhane2021}, define \enquote{diversity} narrowly, and do not account for the long-tail of image categories that are common in other cultures. For example, items and activities common in other parts of the world, such as those in the Arab world, are under-represented, if at all, in existing image databases \cite{Yang2020}. Examples include traditional daily clothing items, such as the \enquote{thobe}, and sporting activities, such as falconry. We refer to these under-represented categories, in which \textit{no} images are available in existing databases, as the \enquote{hidden tail}. This is analogous to the \enquote{long tail} of image categories, in which \textit{few} images are available, that the machine learning community has dedicated substantial effort to better representing. 

% Implications of non-diverse datasets
Such an exclusion of culturally-diverse images has a technical, societal, and ethical impact on the machine learning community. From a technical perspective, the absence of diverse images in existing databases violates the assumption that samples are from \enquote{the wild} and representative of the underlying data distribution. By evaluating networks on such narrow samples, their performance tends to be an over-estimate. Moreover, culturally-diverse image categories are effectively out-of-distribution (OOD) samples notorious for degrading the performance of trained networks \cite{Hendrycks2019Adversarial}, a phenomenon shown to be more prominent when transferring across geographical regions \cite{Hendrycks2020OOD}. On a societal level, pre-trained networks are less likely to be of direct value to researchers residing in, or operating with, under-represented communities. This is driven by the poor performance of such networks on OOD samples, which is a direct consequence of the cultural bias inherent in the datasets used to train such networks. With this imbalance in the applicability of networks across cultures, under-represented communities are unlikely to capture the benefits of computer vision-based advancements. Furthermore, the machine learning community's lack of exposure to data from diverse cultures suggests that researchers have less of an opportunity to learn about such cultures. Such dataset-based learning, the acquisition of skills and knowledge via datasets, has been evident with, for example, the Caltech-UCSD Birds 200 database \cite{Wah2011} and ornithology. On an ethical level, the absence of data to which researchers can relate implicitly excludes these researchers from more actively engaging with the machine learning community. As such, it is to the advantage of the community to build the infrastructure that incentivizes the involvement of practitioners from a more diverse background in machine learning. 

% Our contribution
In this work, we aim to increase the cultural diversity of images that are available for training neural networks. Hence, we present the Turath-150K\footnote{Turath roughly means heritage in Arabic} database, a large-scale dataset of images depicting objects, activities, and scenarios that are rooted in the Arab world and culture. We chose this culture as an exemple, particularly due to its under-representation in existing publicly-available datasets, and hope other researchers follow suit with publishing datasets depicting cultures from around the globe. Specifically, our contributions are the following: (1) we build a large-scale database of images, entitled Turath-150K, the first of its kind that centres around life in the Arab world. For benchmarking purposes, we split the database into three distinct subsets; Turath-Standard, Turath-Art (focusing on art from the Arab world), and Turath-UNESCO (focusing on heritage sites located in the Arab world). (2) We shed light on the limitations of deep neural networks pre-trained on ImageNet by showing that they are unable to deal with the out-of-distribution samples of the Turath database. (3) We evaluate various networks on the Turath benchmark databases and demonstrate their image classification performance on both high and low-level categories. 

\section{Related work}

There exists a multitude of publicly-available image databases that have been exploited for the training of deep neural networks. We outline several that we believe are most similar to our work and also elucidate how our database, Turath, differs significantly in motivation, scope, and content. 

\paragraph{Scene recognition databases}
The task of scene recognition involves identifying scenes based on images. To facilitate achieving this task, the SUN397 database \cite{xiao2010sun} was designed to contain 100K images of 397 scenes. The vast majority of these scene categories are motivated by the WordNet hierarchy \cite{Fellbaum2010}. Similarly, the Places database \cite{zhou2017places} was designed to contain 2.5 million images of 365 high-level scenes, such as coffee-shop, nursery, and train station. Although extensive in terms of the number of samples, the scene categories lack the granularity that we offer and do not trivially extend to the Arab world. Moreover, Turath is not exclusively limited to scenes (see Sec.~\ref{section:construction}) and goes beyond the narrow WordNet hierarchy by explicitly accounting for entities in the Arab world.

\paragraph{Object classification databases}
The task of object classification focuses on identifying object(s) in an image. To propel research on this front, the Caltech 256 database \cite{Griffin2007} was designed to contain 30K images of everyday objects, such as cameras and laptops. The COCO database \cite{Lin2014} is much more extensive with 330K images corresponding to 80 object categories and consisting of multiple annotations, including segmentation maps at various levels of detail. Nonetheless, such databases differ in motivation, scope, and content from our database. In order to increase the cultural diversity of datasets, we turn our attention to objects, activities, and scenarios commonly found in the Arab world. Moreover, our image annotations are not only absent from existing databases but also offer a finer resolution of class label. We explain this in further depth in the next section.  

\paragraph{Out-of-distribution databases}
Researchers have adopted various approaches to handle the generalization of their models to out-of-distribution samples. These approaches can be split according to whether they are implemented during training or evaluation, with the latter being more relevant to our work. For example, ImageNet-R \cite{Hendrycks2020OOD} is an evaluation database of 30K images, spanning 200 ImageNet categories, rendered in different styles and textures. While their approach augments existing ImageNet categories, our database includes image samples from categories \textit{beyond} the ImageNet-1K. ImageNet-O \cite{Hendrycks2019Adversarial} is an evaluation database that claims to reflect label distribution shift, yet still only comprises images from 200 categories in ImageNet-1K. Whereas ImageNet-O is focused on evaluating out-of-distribution detectors, the Turath database is primarily focused on increasing the representation of image categories that are under-represented in ImageNet.

% \color{blue}
% \textbf{TODO - }search for papers explcitly talking about the under-representativeness of ImageNet/similar databases. \url{https://arxiv.org/pdf/1912.07726.pdf} talk about bias in the 'person' synset in ImageNet and quantify this but do not address the bias.
% \color{black}

\section{Design and construction of the Turath database}
\label{section:construction}

In light of our emphasis on increasing the cultural diversity of images, we aimed to construct a database that satisfies the following desiderata:

\begin{enumerate}[leftmargin=0.4cm]
    \item \textbf{Heritage - }Categories of images must be specific to the cultures of the Arab world; we reiterate that although our particular choice of culture stems from its under-representation in existing publicly-available databases, it is simply an example. There remains a multitude of rich cultures that are under-represented and we hope other researchers eventually publish such culture-specific databases, be they in the form of images, audio, or video. 
    \item \textbf{Quantity - }Each category must contain a sufficient number of images to facilitate learning; although the term \enquote{sufficient} is nebulous and category-dependent, existing databases have demonstrated success with at least 50 images per category. We quadruple that amount and aim for at least 200 images per category.
    \item \textbf{Real World - }Images in each category must reflect those commonly encountered \enquote{in the wild}; networks trained on image databases have a number of applications but they are, arguably, most useful when applied in the real world to challenges afflicting stakeholders from patients to farmers. To that end, we aim to collect natural RGB images.
\end{enumerate}

% \color{blue}
% Enumerate the limitations of existing datasets (similar to that in \url{https://ieeexplore.ieee.org/stamp/stamp.jsp?tp=&arnumber=9248596}).
% \color{black}

%\subsection{Construction of database}

The construction of the Turath database consisted of three main stages. We first defined keywords to guide the download of images from web-based search engines. We then used these keywords to assign images an annotation. Lastly, and as a form of noise reduction, we trained several classifiers to distinguish between categories and removed images that were likely to be associated with the incorrect annotation. We now describe these stages in more depth.

\paragraph{Stage 1: Defining keywords and downloading the images}

% For example, ImageNet and COCO exploit the WordNet hierarchy to guide their choice of categories of images. Other databases, such as Places, continue to leverage this hierarchy and introduce various adjectives into their search terms to \nequote{diversify} their images, where diversity refers to the conditions in which images were taken (e.g., night, foggy, etc.). 

Existing image databases such as ImageNet and Places were created by performing query-based searches using online search engines. In this setting, the choice of queries determines the type and quality of images that are retrieved. In our context, and in contrast to the aforementioned work, the WordNet hierarchy \cite{Fellbaum2010} did not satisfy our outlined desiderata. This is primarily because WordNet was not designed for the Arab world and thus does not contain categories that are directly relevant for our purposes. Although an Arabic WordNet \cite{Black2006} does exist, it is unable to capture the cultural focus and the \textit{micro} categories (described next) that we are searching for. 

\begin{figure}[!t]
    \centering
    % \begin{subfigure}{0.48\textwidth}
    % \includegraphics[width=\textwidth]{Standard_Macro_A.png}
    % \end{subfigure}
    % \hfill
    % \begin{subfigure}{0.48\textwidth}
    % \includegraphics[width=\textwidth]{Standard_Macro_B.png}
    % \end{subfigure}
    % ~
    \begin{subfigure}{0.32\textwidth}
        \centering
        \includegraphics[width=\textwidth]{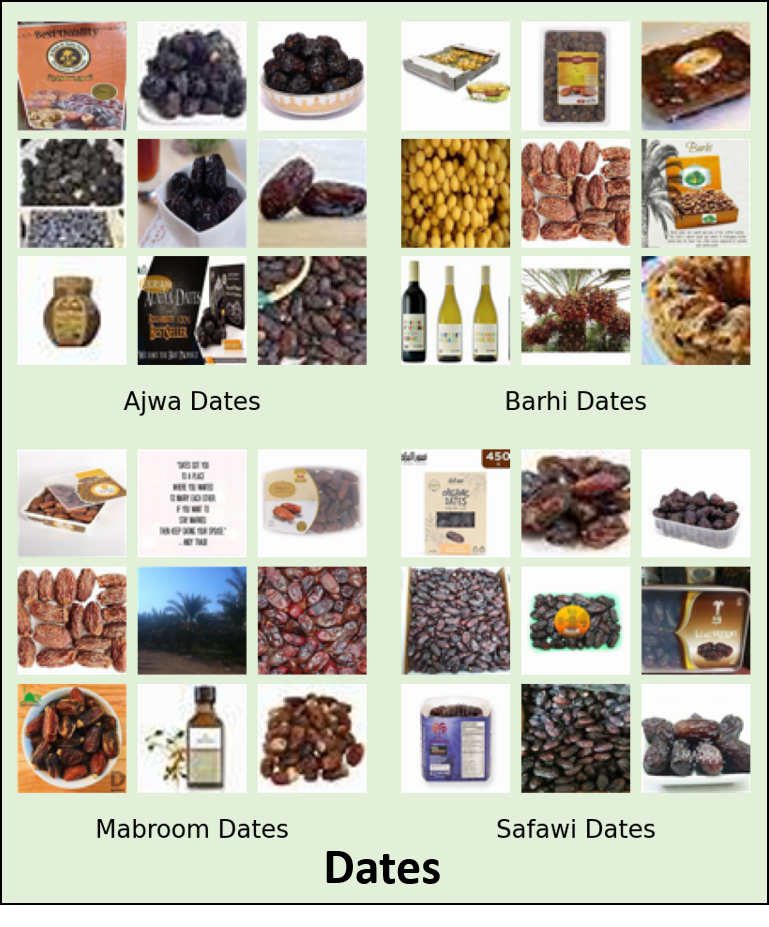}
    \end{subfigure}
    \hfill
    \begin{subfigure}{0.32\textwidth}
        \centering
        \includegraphics[width=\textwidth]{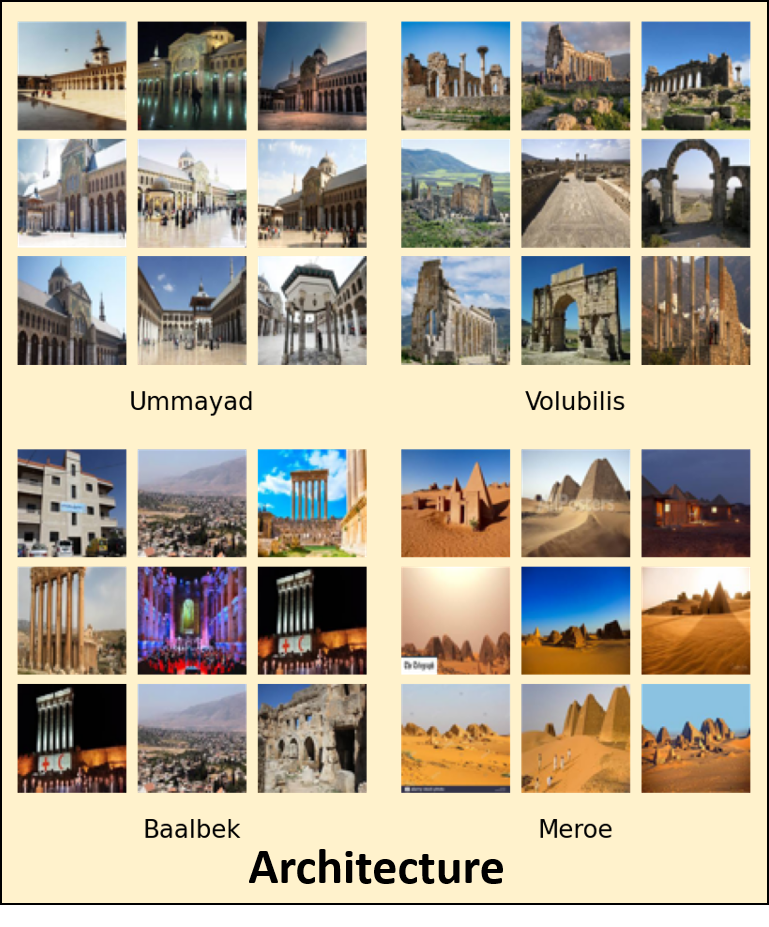}
    \end{subfigure}
    \hfill
    \begin{subfigure}{0.32\textwidth}
        \centering
        \includegraphics[width=\textwidth]{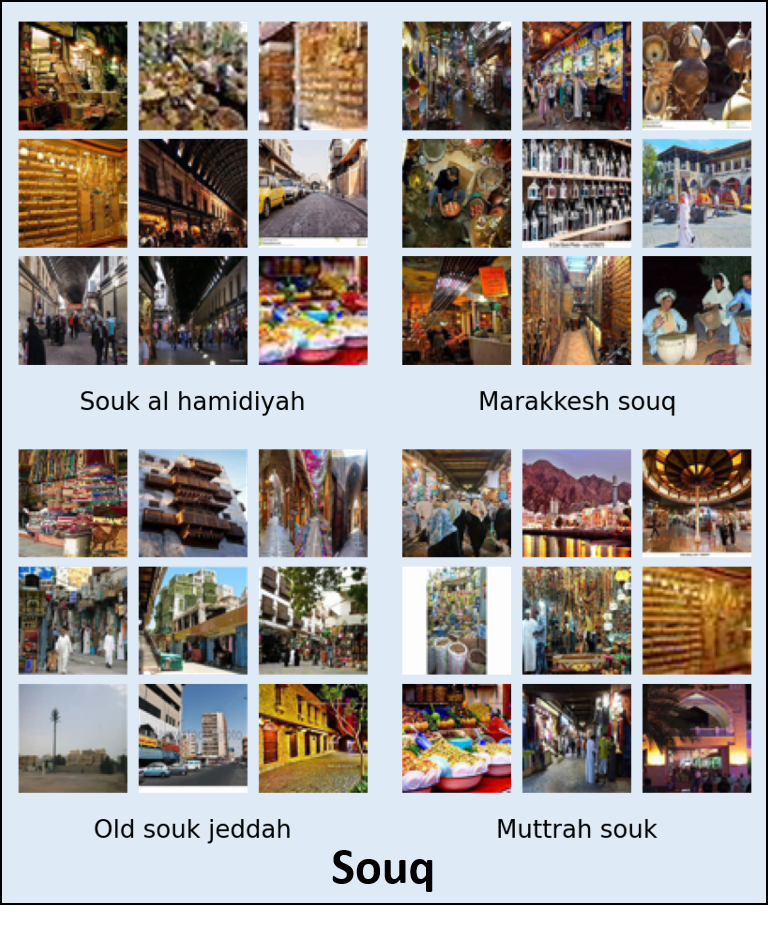}
    \end{subfigure}
    \caption{\textbf{Images samples from a subset of categories available in Turath.} Four \textit{micro} categories are shown for each of the three \textit{macro} categories, $\mathrm{Dates}$, $\mathrm{Architecture}$, and $\mathrm{Souq}$. The image categories range from objects with low-level details, such as dates, to locations with high-level details, such as architecture.}
    \label{fig:images}
\end{figure}

Given our emphasis on the Arab world as an example, we conducted query-based searches of entities engrossed in the diverse cultures of the region. This ranged from categories of images with a low level of detail, such as cities and architecture, to those with a high level of detail, such as traditional food and clothing. Each of these \textit{macro} categories are formed by grouping several \textit{micro} categories. For example, the \textit{macro} category of $\mathrm{Cities}$ comprises $25+$ \textit{micro} categories of images from specific cities in the Arab world, e.g., Damascus, Cairo, and Casablanca. To emphasize the under-representation of images of these cities in existing databases, we note that the largest image database of cities, World Cities \cite{Tolias2011}, with 2.25M images, covers a single city (Dubai) in the Arab world. In Fig.~\ref{fig:images}, we present image samples from three macro categories, $\mathrm{Dates}$, $\mathrm{Architecture}$, and $\mathrm{Souq}$, each containing four \textit{micro} categories.

In addition to retrieving images from the categories mentioned above, we dedicate time and effort to curating two additional \textit{macro} categories that comprise a large number of \textit{micro} categories. Specifically, these revolve around Arab Art and United Nations Educational, Scientific and Cultural Organization (UNESCO) sites. When retrieving images that belong to the Arab Art category, we followed the same strategy of query-based searches. However, given the breadth of this field and to keep the task of downloading images tractable and organized, our search queries were based on artists' names. To that end, we identified 425 names available on the Barjeel Art Foundation website\footnote{\url{https://www.barjeelartfoundation.org/}}. As for the UNESCO category, our search queries were based on the names of 88 recognized UNESCO sites in the Arab world\footnote{\url{https://whc.unesco.org/en/list/&&&order=region}}. 

\paragraph{Stage 2: Labelling the images using keywords}

Each image in the Turath database has two image-level annotations; a \textit{micro} label and a \textit{macro} label. To assign downloaded images to \textit{micro} categories, we follow the strategy proposed by Marin \textit{et al.} \cite{Marin2019} where each category is defined by the query used to search for those images. Similar to their conclusions, we also find that such an approach leads to relatively high quality images that are relevant to the search query. We then grouped \textit{micro} categories with similar themes into \textit{macro} categories. As an example, we grouped seven types of dates (\textit{micro}) into a single $\mathrm{Dates}$ category (\textit{macro}).

\paragraph{Stage 3: Filtering the images with classifier-based labelling}

Despite our effort to conduct searches using queries that are unambiguous and descriptive, upon further inspection, we found that certain categories contained images that were irrelevant. This was most prominent amongst images that belonged to artists. For example, the query $\mathrm{inji \ efflatoun \ art}$ returned art pieces associated with the artist $\mathrm{Inji \ Efflatoun}$, as desired, but also images of the artist herself. 

\begin{figure}[!t]
    \centering
    \begin{subfigure}{0.55\textwidth}
    \centering
    \includegraphics[width=\textwidth]{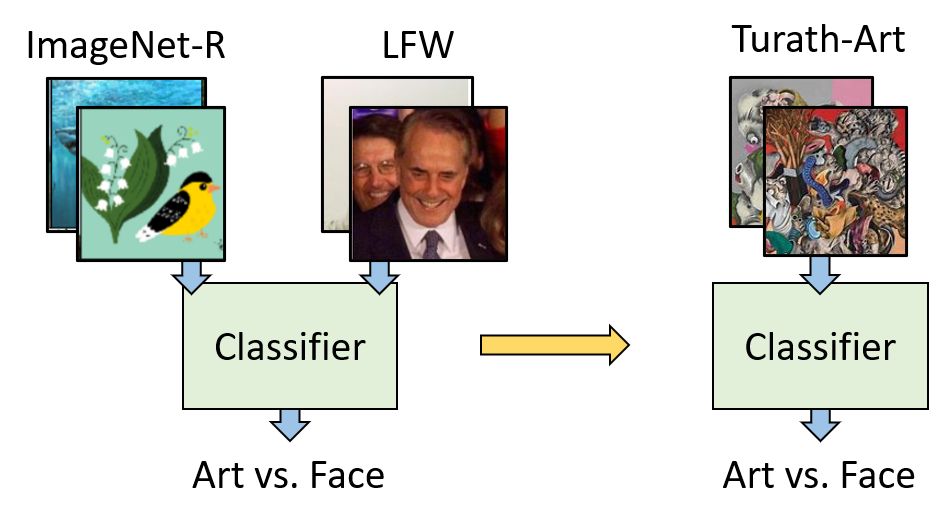}
    \end{subfigure}
    ~
    \begin{subfigure}{0.4\textwidth}
    \centering
    \includegraphics[width=1\textwidth]{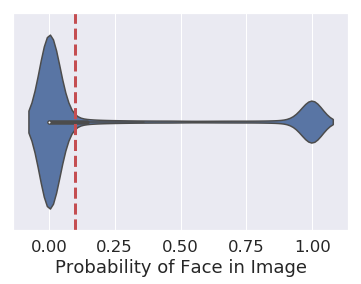}
    \end{subfigure}
    \caption{\textbf{Pipeline for cleaning data in Turath database.} \textbf{(Left)} Classifier-based cleaning of data. We trained a binary classifier to distinguish between images of art (ImageNet-R) and faces (LFW) and deployed it on Turath-Art. \textbf{(Right)} Distribution of probabilities output by binary classifier deployed on all images of Turath Art. We found that, when a threshold of $0.1$ is chosen, approximately 26.1\% of images are identified as a face.}
    \label{fig:art_vs_faces}
\end{figure}

To remedy this situation, we exploited the prior knowledge that out-of-distribution (OOD) image samples are likely to be of artists' faces. Therefore, given our emphasis on retaining images of art pieces, we designed a binary classifier that distinguished between images of art and those of faces. To train such a classifier, we needed images with relatively high quality labels. For those in the \enquote{art} domain, we grouped all the categories in ImageNet-R \cite{Hendrycks2020OOD}, which comprises images from ImageNet rendered artistically, into a single category. For those in the \enquote{faces} domain, we exploited images from the LFW database \cite{LFW2007}, which comprises 13K images of faces, and grouped them into a single category. After training this classifier, we performed inference on \textit{our} set of artistic images. Given that the majority of images are those of art pieces, we would expect the distribution of output probabilities to be bi-modal and skewed towards the value zero (i.e., corresponding to art images). This is indeed what we find empirically, as shown in Fig.~\ref{fig:art_vs_faces}. Upon manual inspection of the images, we chose a threshold value of $0.1$, whereby approximately 26.1\% of image samples believed to have been of art are instead identified as a face. These 27,302 images are removed from the database. 

% We first trained a network on the ImageNet-R dataset \cite{Hendrycks2020OOD}, which comprises images from ImageNet rendered artistically. We then exploited an established OOD detection method \cite{Hendrycks2016Baseline} where, after we performed a forward pass, using the trained network, on all images in \textit{our} set of artistic images, we calculated the (negative) maximum softmax probability (MSP) as a proxy for an anomalous image sample. However, this process offers minimal insight when dealing with large-scale databases because it requires the non-trivial act of defining a threshold on the MSP. Therefore, we turned our attention to a different OOD method.

Detecting OOD images of human faces exploited the implicit bias that human faces comprised the majority of the OOD images. However, not all OOD images contain human faces. To investigate this, we explored more general approaches involving one-class SVMs \cite{erfani2016high}, deep autoencoding GMMs \cite{zong2018deep}, adversarial networks \cite{li2018anomaly}, geometric transformations \cite{golan2018deep} and self-supervised classification networks \cite{amrani2021self}. We empirically found that although this self-supervised approach was preferable to the remaining methods, it was still unable to reliably identify OOD samples.

% To remedy this situation, we took several steps. First, we trained a binary classifier to distinguish between images within and outside of the Art category. We then performed inference on all of the images in the Art category to identify those which were most confidently labelled as Art. Such images with high-confidence labels were kept and the remaining were discarded. 

% \color{blue}
%  How about ensembles? We train multiple OOD detectors and take majority score. 
% \color{black}

% \color{blue}
% OPTION2 - we can follow the work of \url{https://arxiv.org/abs/1802.04865}
% \begin{enumerate}
%     \item Extra head that predicts scalar confidence value per sample
%     \item Threshold applied to confidence value can distinguish between in and out-of distribution samples (e.g., confidence < $\delta$ = OOD)
%     \item We can find/calibrate $\delta$ by using misclassified validation samples as a 'proxy' for true OOD samples
% \end{enumerate}
% \color{black}

\section{Turath benchmark databases}
\label{section:benchmarks}

The Turath database comprises three specialized subsets of data that contain images from mutually-exclusive categories. Hereafter, these subsets will be referred to as Turath Standard, Turath Art, and Turath UNESCO, respectively, and, in this section, will be described in depth. We chose to separate the database along these dimensions to account for the different resolution of the categories, as will be shown next. 

% \color{blue}
% We can calculate the Frechet Inception Distance (FID) between images \textit{across} benchmarks (Standard vs. Art. vs. UNESCO) to substantiate our claim that they contain images from distinct categories i.e., there is minimal similarity between these images.
% \color{black}

\paragraph{Turath Standard}

\begin{figure}[!h]
    \vspace{-\intextsep}
    \centering
    \begin{subfigure}{0.48\textwidth}
    \centering
    \includegraphics[width=1\textwidth]{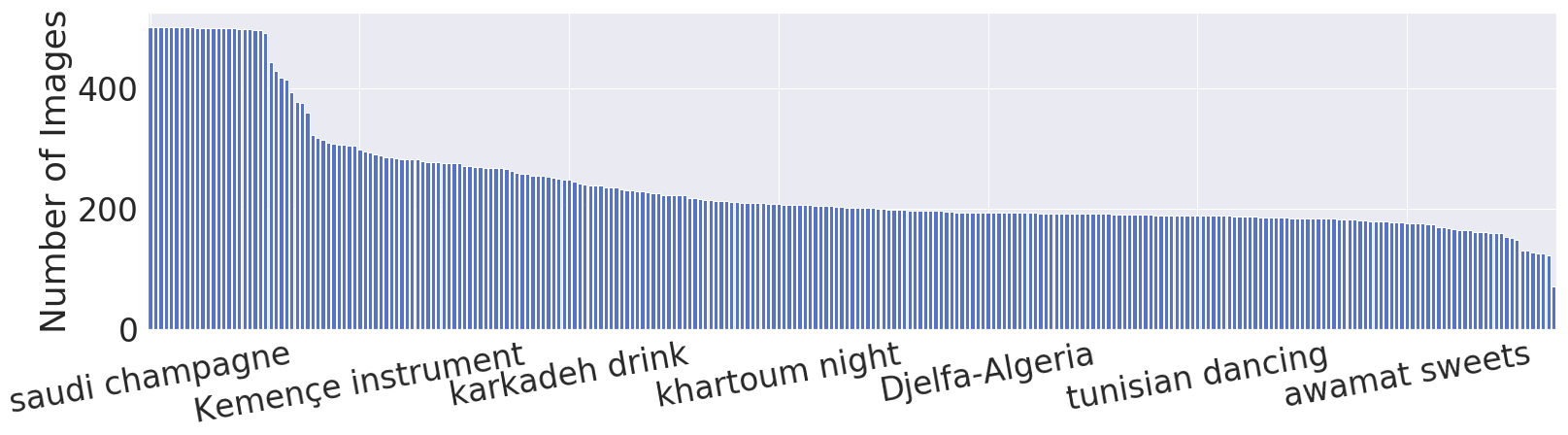}
    \caption{Turath Standard}
    \label{fig:histogram_standard}
    \end{subfigure}
    ~
    \begin{subfigure}{0.48\textwidth}
    \centering
    \includegraphics[width=1\textwidth]{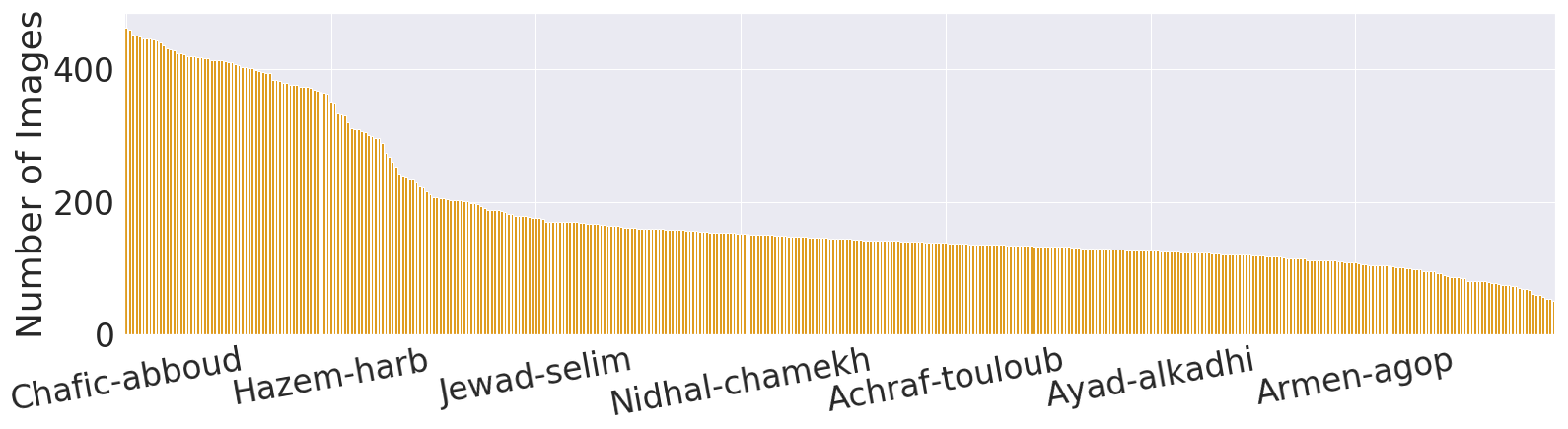}
    \caption{Turath Art}
    \label{fig:histogram_art}
    \end{subfigure}
    ~
    \begin{subfigure}{0.48\textwidth}
    \centering
    \includegraphics[width=1\textwidth]{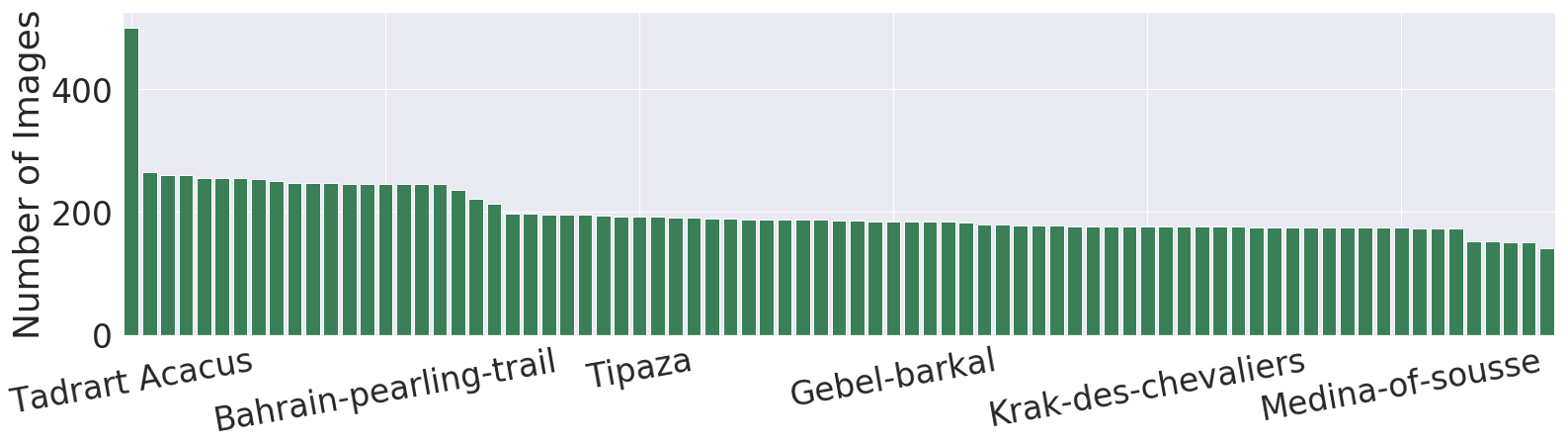}
    \caption{Turath UNESCO}
    \label{fig:histogram_unesco}
    \end{subfigure}
    \caption{\textbf{Number of images per \textit{micro} category in each of the benchmark databases.} Each micro category contains anywhere between 50-500 images. For clarity, we present only a subset of the micro category names. The full list of categories can be found in Appendix~\ref{appendix:database_categories}.}
    \label{fig:dataset_summary}
\end{figure}

The Turath Standard benchmark database comprises images reflecting the diverse range of objects, activities, and scenarios commonly encountered in the Arab world. Each image has a \textit{macro} and \textit{micro} image-level category annotation. The twelve macro categories are $\mathrm{Cities}$, $\mathrm{Food}$, $\mathrm{Nature}$, $\mathrm{Architecture}$, $\mathrm{Dessert}$, $\mathrm{Clothing}$, $\mathrm{Instruments}$, $\mathrm{Activities}$, $\mathrm{Drinks}$, $\mathrm{Souq}$, $\mathrm{Dates}$, and $\mathrm{Religious \ Sites}$. The complete list of the more granular \textit{micro} categories can be found in Appendix~\ref{appendix:database_categories}. The number of images in each of these micro categories is presented in Fig.~\ref{fig:histogram_standard}. We can see that each micro category has anywhere between $50-500$ images. This is by design since we explicitly searched for \textit{up to} 500 images per category and excluded categories with fewer than 50 images. We applied this strategy to all benchmark databases to avoid categories with too few images which may contain noise and thus hinder a network's ability to learn.

\begin{wraptable}[11]{l}{0.6\textwidth}
\small
\vspace{-\intextsep}
\centering
\caption{\textbf{Overview of training, validation, and test splits for the Turath benchmark databases.} The number of macro categories is shown in brackets.}
\label{table:data_splits}
\begin{tabular}{c|c c c}
     \toprule
     & \multicolumn{3}{c}{Turath Database} \\
     & Standard & Art & UNESCO \\
     \midrule
     Training & 38,894 & 46,665 & 9,540 \\
     Valid. & 6,418 & 7,531 & 1,558 \\
     Test & 19,472 & 22,969 & 4,778 \\
     \midrule
     Categories & 269 (12) & 419 & 79 \\
     \bottomrule
\end{tabular}
\end{wraptable}

For benchmarking, the Turath Standard database contains 38,894 images in the training set, 6,418 images in the validation set, and 19,472 images in the test set (see Table~\ref{table:data_splits}). Unless otherwise specified, all data splits are performed uniformly at random with a ratio of 70:10:20 for the training, validation, and test sets, respectively.

% \begin{wrapfigure}[25]{l}{0.6\textwidth}
% \end{wrapfigure}

\paragraph{Turath Art} 

The Turath Art benchmark comprises images of art (e.g., paintings, sculptures, etc.) created by Arab artists alongside annotations, at the image-level, of such artists. We purposefully excluded these categories from the Turath Standard benchmark for the following reasons. First, the large number of \textit{micro} categories ($419$) that would have fallen under the \textit{macro} category of $\mathrm{Art}$ would have overwhelmed the categories outlined in the Turath Standard benchmark. Second, distinguishing between images containing intricate, low-level details reflected by paintings, sculptures, etc., poses a difficult task, in and of itself. As a result, this warranted a distinct specialized benchmark, which we refer to as Turath Art. In Fig.~\ref{fig:histogram_art}, we present the number of images in each of the 419 artist categories, and include a subset of the artists' names for clarity. For benchmarking, the Turath Art database contains 38,445 images in the training set, 6,354 images in the validation set, and 19,324 images in the test set.

\paragraph{Turath UNESCO}

The Turath UNESCO benchmark comprises images of UNESCO world heritage sites in the Arab world alongside annotations, at the image-level, of these sites. We present, in Fig.~\ref{fig:histogram_unesco}, the total number of images in each of the 79 categories. For benchmarking, the Turath UNESCO database contains 9,540 images in the training set, 1,558 images in the validation set, and 4,778 images in the test set.

% \begin{table}[!h]
%     \centering
%     \caption{Number of images in the training, validation, and test sets of the Turath benchmark databases alongside the number of categories in each benchmark.}
%     \label{table:number_of_images}
%     \begin{tabular}{c|c c c c}
%          \toprule
%          & Training & Valid. & Test & Categ. \\
%          \midrule
%          Standard & 38,894 & 6,418 & 19,472 & 12 \\
%          Art & 46,665 & 7,531 & 22,969 & 419 \\
%          UNESCO & 9,540 & 1,558 & 4,778 & 79\\
%          \bottomrule
%     \end{tabular}
% \end{table}

\section{Experimental results}
\label{section:results}

\subsection{Limitations of networks pre-trained on ImageNet}

\begin{figure}[!h]
    \centering
    \begin{subfigure}{1.0\textwidth}
        \includegraphics[width=\textwidth]{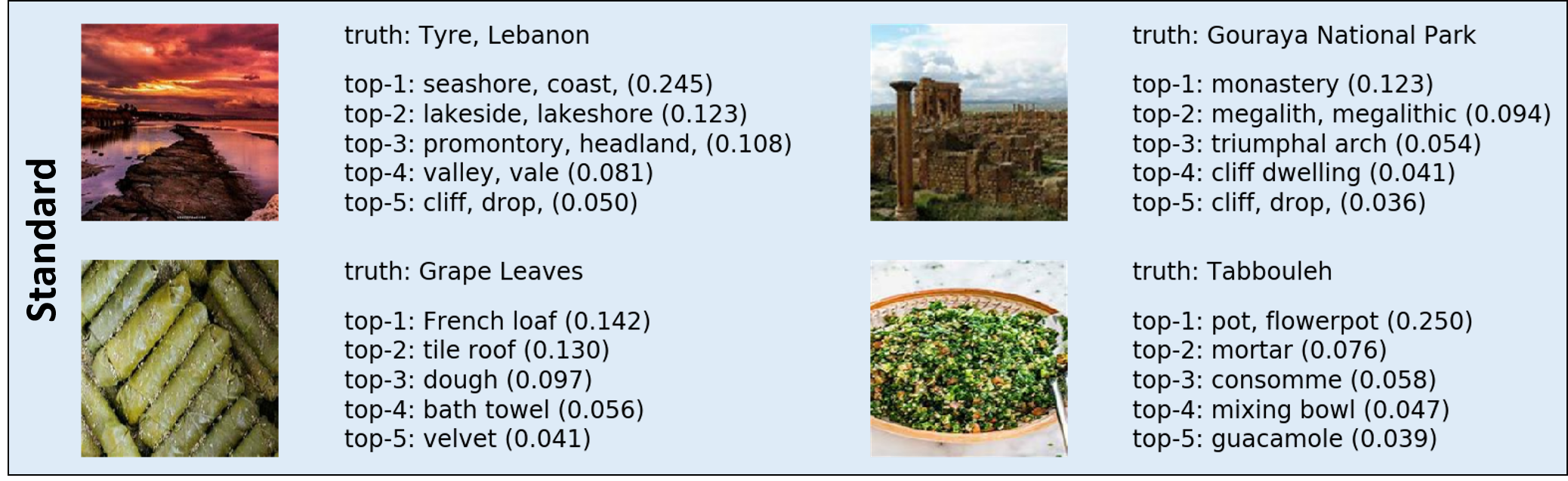}
        % \caption{Turath Standard}
        % \label{fig:failures_standard}
    \end{subfigure}
    ~
    \begin{subfigure}{1.0\textwidth}
        \includegraphics[width=\textwidth]{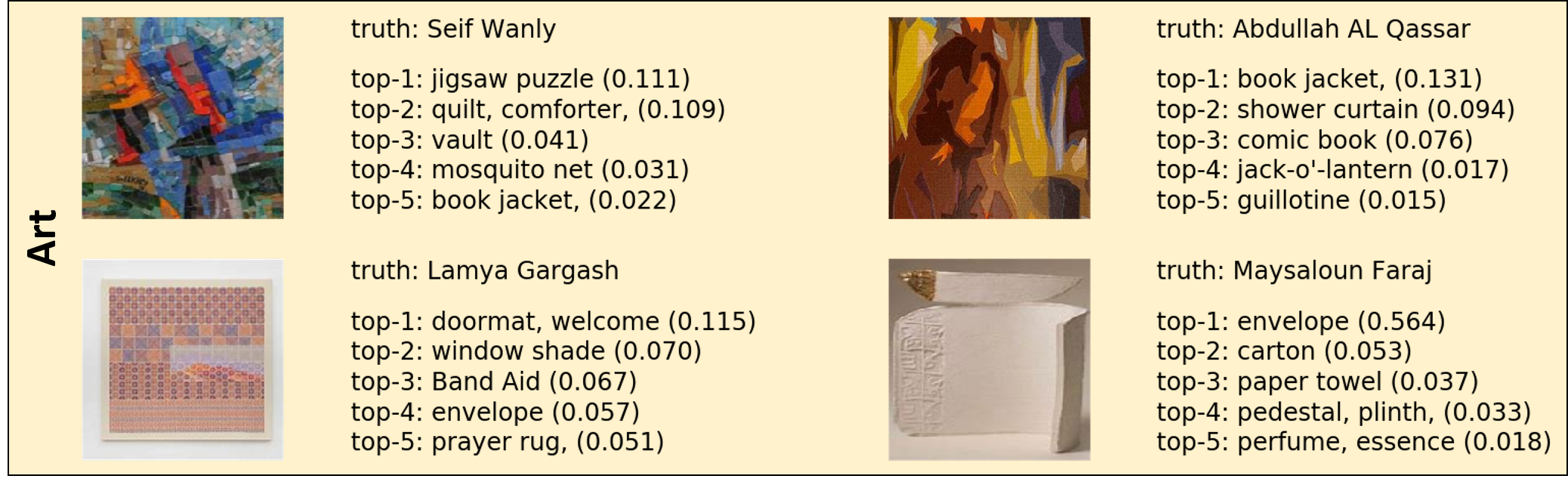}
        % \caption{Turath Art}
        % \label{fig:failures_art}
    \end{subfigure}
    ~
    \begin{subfigure}{1.0\textwidth}
        \includegraphics[width=\textwidth]{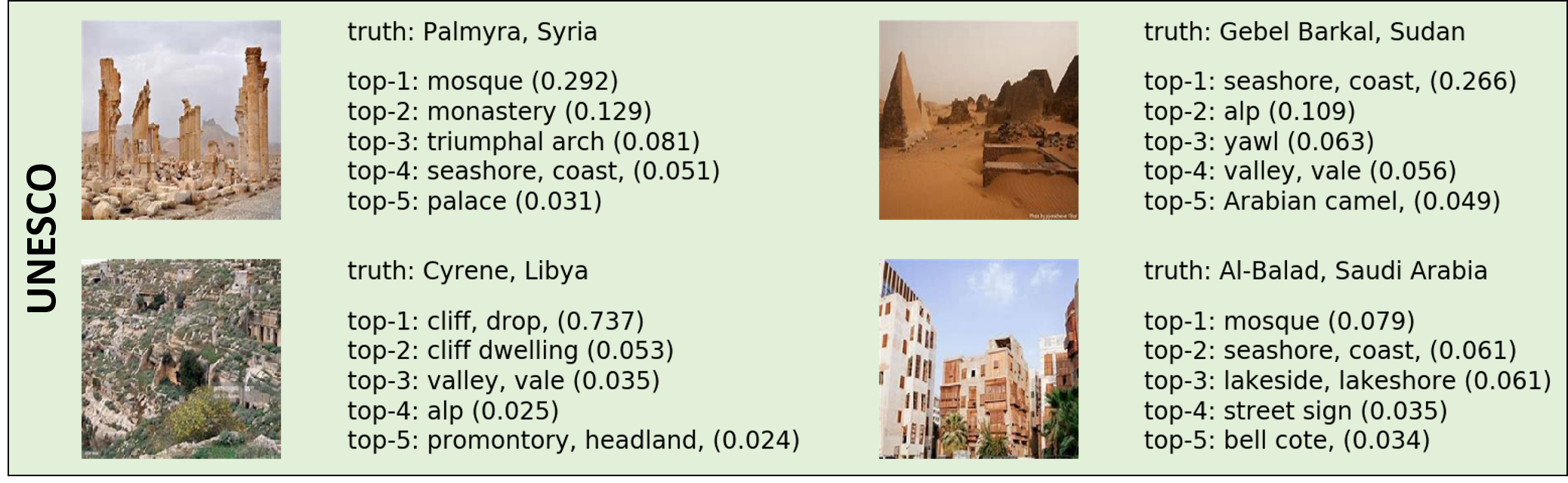}
        % \caption{Turath UNESCO}
        % \label{fig:failures_unesco}
    \end{subfigure}
    \caption{\textbf{Top-5 predictions (and confidence) made by an EfficientNet, pre-trained on ImageNet and directly deployed on image samples from the Turath benchmark databases.} We also present the ground-truth \textit{micro} category of each of the image samples. Many of the predictions assign a high probability mass to the incorrect category, lack the finer resolution of our \textit{micro} categories, and do not have a cultural emphasis.}
    \label{fig:efficientnet_failures}
\end{figure}

The utility of a pre-trained neural network is contingent upon the similarity of the upstream task, on which the network was trained, and the downstream task, on which the network is deployed \cite{Raghu2019}. To qualitatively evaluate this utility in the context of the Turath database, we randomly sample images from each of the benchmark databases, perform a forward pass through an EfficientNet \cite{Tan2019} pre-trained on ImageNet, and compare the Top-5 predictions to the ground-truth label (see Fig.~\ref{fig:efficientnet_failures}). We find that, across the benchmarks, EfficientNet assigns a high probability mass to incorrect image categories. For example, it classified a sculpture by the artist $\mathrm{Maysaloun \ Faraj}$ as an envelope with a confidence score ($0.564$) and $\mathrm{Gebel \ Barkal}$, pyramids in Sudan, as a $\mathrm{seashore}$ with a confidence score ($0.266$). These results also suggest that confidence-based decisions, such as network classification abstention and out-of-distribution detection \cite{Hendrycks2016Baseline}, may be of little value in this context. We show that these limitations also extend to other neural architectures (see Appendix~\ref{appendix:network_limitations}). 

\subsection{Image classification on Turath benchmark databases}

In this section, we adapt networks pre-trained on ImageNet using data from the Turath database benchmarks. We do so by introducing, and randomly initializing, a classification head, $p_{\theta}: h \rightarrow \hat{y} \in \mathbb{R}^{C}$, that maps the penultimate representation, $h$, of the feature extractor network to the predicted probability distribution, $\hat{y}$, over the set of image categories, $C \in \{12,269,419,79\}$ depending on the benchmark database. In the linear evaluation phase, we freeze the parameters of the feature extractor network whereas in the fine-tuning phase, we use those parameters as an initialization and update them accordingly. In both phases, we train networks using the Adam optimizer with a categorical cross-entropy loss and a learning rate, $lr \in [1e^{-3}, 1e^{-4}]$. Further implementation details can be found in Appendix~\ref{appendix:implementation_details}. 

In Table~\ref{table:results}, we present the Top-1 and Top-5 accuracy achieved by networks in these experiments. The Top-1 accuracy refers to the percentage of image samples whose ground-truth category matches the category most confidently predicted by the network. In contrast, Top-5 accuracy refers to the percentage of images samples whose ground-truth category can be found in the Top-5 most confident predictions made by the network\footnote{We provide demos of these networks in action at \url{danikiyasseh.github.io/Turath/[benchmark]Demo} where benchmark $\in$ [Standard, Art, UNESCO].}. On average, we find that EfficientNet outperforms MobileNetV2 and ResNet50 uniformly across the benchmark databases. For example, on the UNESCO database, EfficientNet, in the linear evaluation phase, achieves Top-1$= 39.5$ whereas MobileNetV2 and ResNet50 achieve Top-1$= 32.1$ and $33.2$, respectively. We also show that the \textit{micro} category image classification tasks across benchmark databases differ in their level of difficulty. This is evident by the large range of reported accuracy scores. For example, Turath Standard poses the least difficult task with a best Top-1$=46.1$ whereas Turath Art poses the most challenging task with a best Top-1$=16.5$. This is expected given the high similarity of images in the Art database. We believe these accuracy scores, which remain relatively lower than those achieved on ImageNet (Top-1=$90.2$), stand to benefit from further advancements in neural architecture design, transfer learning, and domain adaptation. We also find that fine-tuning networks, regardless of the architecture, is more advantageous than a linear evaluation of such networks. This suggests that the fixed features extracted from a network pre-trained on ImageNet are relatively constraining.

% \begin{table}[!h]
% \begin{wraptable}[23]{L}{0.55\textwidth}
%     \centering
%     \caption{Image classification accuracy on the Turath Standard (macro), Art, and UNESCO benchmark databases. Results are averaged across five random seeds. \color{blue} Experiments underway \color{black}}
%     \label{table:results_standard}
%     \begin{tabular}{l c c c c}
%         \toprule
%          & \multicolumn{2}{c}{Validation Set} & \multicolumn{2}{c}{Test Set} \\
%          \cmidrule{2-5} 
%          Architecture & Top-1 & Top-5 & Top-1 & Top-5 \\
%          \midrule
%          \multicolumn{5}{l}{\textit{Turath Standard}} \\
%          \midrule
%          MobileNetV2 & 71.1 & 97.2 & 70.1 & 96.8 \\
%          EfficientNet & \textbf{74.7} & \textbf{97.6} & \textbf{71.1} & 96.6 \\
%          ResNet50 & 70.5 & 97.1 & 69.7 & 96.9 \\
%          \midrule
%          \multicolumn{5}{l}{\textit{Turath Art}} \\
%          \midrule
%          MobileNetV2 &  \\
%          EfficientNet &  \\
%          ResNet50 &  \\
%          \midrule
%          \multicolumn{5}{l}{\textit{Turath UNESCO}} \\
%          \midrule
%          MobileNetV2 & 56.4 & 76.2 & 32.1 & 53.6 \\
%          EfficientNet & \textbf{58.5} & \textbf{78.0} & \textbf{39.5} & \textbf{60.6}\\
%          ResNet50 & 61.5 & 77.8 & 33.2 & 54.0 \\
%          \bottomrule
%     \end{tabular}
% \end{wraptable}

\begin{table}[!t]
    \centering
    \caption{\textbf{Image classification test accuracy on the Turath Standard, Art, and UNESCO benchmark databases.} Results are averaged across five random seeds and standard deviation is shown in brackets. Bold results reflect the best-performing network architecture in each benchmark.}
    \label{table:results}
    \resizebox{\textwidth}{!}{
    \begin{tabular}{l c c c c c c c c}
        \toprule 
         & \multicolumn{2}{c}{Standard \textit{(macro)}} & \multicolumn{2}{c}{Standard \textit{(micro)}} & \multicolumn{2}{c}{Art} & \multicolumn{2}{c}{UNESCO} \\
         Architecture & Top-1 & Top-5 & Top-1 & Top-5 & Top-1 & Top-5 & Top-1 & Top-5 \\
        %  \midrule
        %  \multicolumn{9}{l}{\textit{Random initialization \color{blue} underway \color{black}}} \\
        %  \midrule
        %  MobileNetV2 & \\
        %  EfficientNet & \\
        %  ResNet50 &  \\
         \midrule
         \multicolumn{9}{l}{\textit{Linear evaluation}} \\
         \midrule
         MobileNetV2 & 70.1 {\scriptsize(0.7)} & 96.8 {\scriptsize(0.1)} & 39.1 {\scriptsize(0.1)} & 62.6 {\scriptsize(0.1)}& 12.7 {\scriptsize(0.2)} & 22.4 {\scriptsize(0.2)} & 32.1 {\scriptsize(0.4)} & 53.6 {\scriptsize(0.2)} \\
         EfficientNet & \textbf{71.2} {\scriptsize(0.3)} & 96.6 {\scriptsize(0.1)} & \textbf{46.1} {\scriptsize(0.2)} & \textbf{69.5} {\scriptsize(0.1)} & \textbf{16.5} {\scriptsize(0.3)} & \textbf{25.2} {\scriptsize(0.3)} & \textbf{39.5} {\scriptsize(0.4)} & \textbf{60.6} {\scriptsize(0.2)} \\
         ResNet50 & 69.7 {\scriptsize(0.2)} & 96.9 {\scriptsize(0.2)} & 39.6 {\scriptsize(0.5)} & 63.4 {\scriptsize(0.3)} & 13.2 {\scriptsize(0.2)} & 23.2 {\scriptsize(0.3)} & 33.2 {\scriptsize(0.3)} & 54.0 {\scriptsize(0.2)} \\
         \midrule
         \multicolumn{9}{l}{\textit{Fine-tuning}} \\
         \midrule
         MobileNetV2 & 65.6 {\scriptsize(1.9)} & 95.6 {\scriptsize(0.3)}  & 41.7 {\scriptsize(1.2)} & 65.9 {\scriptsize(1.3)} & 12.9 {\scriptsize(0.6)} & 23.6 {\scriptsize(0.6)} & 34.4 {\scriptsize(0.7)} & 56.1 {\scriptsize(0.7)} \\
         EfficientNet & \textbf{77.2} {\scriptsize(0.6)} & \textbf{97.6} {\scriptsize(0.0)}  & \textbf{49.9} {\scriptsize(0.3)} & \textbf{73.8} {\scriptsize(0.3)} & \textbf{19.0} {\scriptsize(0.3)} & \textbf{31.2} {\scriptsize(0.4)} & \textbf{43.2} {\scriptsize(0.4)} & \textbf{64.2} {\scriptsize(0.7)} \\
         ResNet50 & 71.4 {\scriptsize(0.7)} & 96.8 {\scriptsize(0.1)}  & 41.2 {\scriptsize(1.3)} & 65.9 {\scriptsize(1.0)} & 14.2 {\scriptsize(0.8)} & 25.0 {\scriptsize(1.1)} & 35.7 {\scriptsize(1.7)} & 56.7 {\scriptsize(1.4)}\\         
         \bottomrule
    \end{tabular}
    }
\end{table}

\begin{figure}[!b]
    \centering
    \begin{subfigure}{0.45\textwidth}
    \includegraphics[width=1\textwidth]{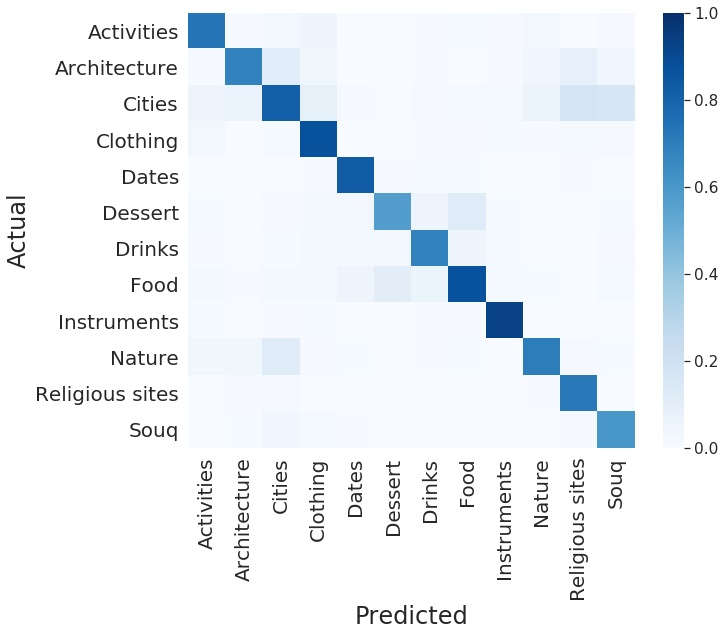}
    \end{subfigure}
    ~
    \begin{subfigure}{0.40\textwidth}
    \includegraphics[width=1\textwidth]{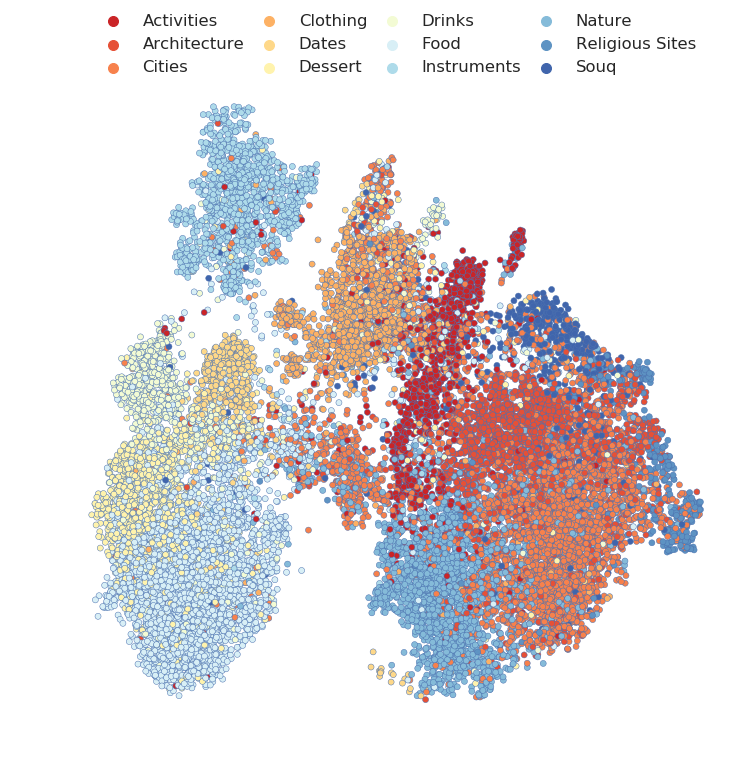}
    \end{subfigure}
    \caption{\textbf{Performance of EfficientNet fine-tuned on the Turath Standard benchmark database.} \textbf{(Left)} Confusion matrix of predictions made on the test set of the Turath Standard benchmark database. Normalization is performed across columns. \textbf{(Right)} UMAP embedding of the penultimate layer representations ($\mathbb{R}^{640}$) of image samples in the test set. We find that the representations exhibit a high degree of separability amongst the macro categories.}
    \label{fig:standard_confusion_matrix}
\end{figure}

To gain better insight on the type of misclassifications committed on Turath Standard, we present, in Fig.~\ref{fig:standard_confusion_matrix} (left), the confusion matrix of macro-category predictions made by EfficientNet on image samples in the test set of the Turath Standard benchmark. This is complemented by Fig.~\ref{fig:standard_confusion_matrix} (right) in which we illustrate the UMAP embedding of the penultimate representations ($\mathbb{R}^{640}$) of the same set of image samples. We chose the fine-tuned EfficientNet for these visualizations given its superior performance (see Table~\ref{table:results}). In light of Fig.~\ref{fig:standard_confusion_matrix}, we find that the network is capable of comfortably distinguishing between macro categories. This is evident by the relatively darker diagonal elements in the confusion matrix and the high degree of category-specific separability of the UMAP embeddings. On the other hand, we find that images in the $\mathrm{Food}$ category are occasionally misclassified as $\mathrm{Dessert}$, an error which makes sense given the semantic proximity of these categories.

\begin{figure}[!t]
    \centering
    \begin{subfigure}{0.3\textwidth}
    \includegraphics[width=1\textwidth]{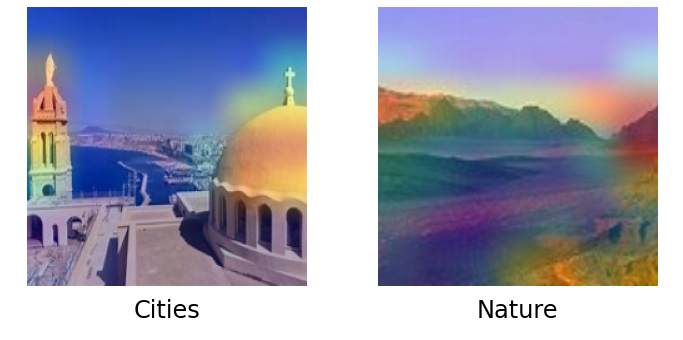}
    \caption{Turath Standard}
    \label{fig:standard_grad_cam}
    \end{subfigure}
    ~
        \begin{subfigure}{0.3\textwidth}
    \includegraphics[width=1\textwidth]{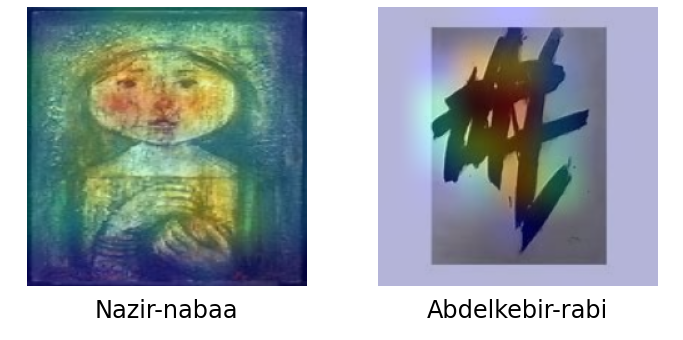}
    \caption{Turath Art}
    \label{fig:art_grad_cam}
    \end{subfigure}
    ~
    \begin{subfigure}{0.3\textwidth}
    \includegraphics[width=1\textwidth]{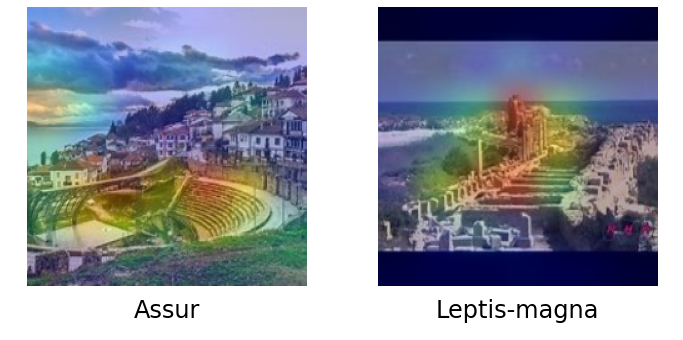}
    \caption{Turath UNESCO}
    \label{fig:unesco_grad_cam}
    \end{subfigure}
    ~
    \caption{\textbf{Heatmap of the most pertinent regions of the image for the category prediction.} We used Grad-CAM with an EfficientNet trained on the Turath (a) Standard, (b) Art, or (c) UNESCO benchmark databases. Red and blue regions are of high and low importance, respectively. We see that the network is able to identify regions in the image appropriate to the image category.}
    \label{fig:grad_cam}
\end{figure}

Having shown that an EfficientNet can adequately learn to distinguish between the various categories in the Turath benchmark databases, we wanted to explore whether its classifications were inferred from the appropriate components of the input image. To do so, we exploit an established deep neural network interpretability method, Grad-CAM \cite{Selvaraju2017}, which attempts to identify the salient regions of the input image in the form of a heatmap. Even though saliency methods have come under scrutiny \cite{Tomsett2020}, we find that, in practice, they can be insightful. In Fig.~\ref{fig:grad_cam}, we illustrate the Grad-CAM-derived heatmap overlaid on the original input image presented to a trained EfficientNet alongside the ground-truth annotation of the image. In the case of $\mathrm{Leptis \ Magna}$ (Fig.~\ref{fig:unesco_grad_cam}), we see that the ancient Carthaginian arches are appropriately identified. 

% Pre-train with ImageNet
% Pre-train with Ours
% Pre-train with ImageNet and Ours
% Evaluate on Western, Evaluate on Ours, Evaluate on Both 

% Train on ImageNet and Turath Standard (expanded classification head)
% Evaluate on Turath Validation Images
% Evaluation on ImageNet Validation Images

\section{Discussion}
\label{section:discussion}

In this paper, we discussed how existing image databases under-represent objects, activities, and scenarios commonly found in certain cultures. To increase the cultural diversity of image databases, we introduced Turath, a database of approximately 150K images of Arab heritage. Moreover, we proposed three specialized benchmark databases, Turath Standard, Art, and UNESCO, that reflect a range of entities within the Arab world and evaluated several deep networks on such benchmarks. Of the networks evaluated, we found that EfficientNet performed best achieving Top-1 accuracy of $49.9$, $19.0$, and $43.2$, on Turath Standard, Art, and UNESCO, respectively. We hope that our benchmark databases can spur the research community to further advance neural architecture design, transfer learning, and domain adaptation. That being said, it is vital that we consider the limitations and broader societal impact of our work.

\paragraph{Limitations} 

When searching for and cleaning the data, we opted out of a crowd-sourcing approach (e.g., Mechanical Turk) in order to scale the database with minimal cost. The machine learning community stands to benefit from the challenge of more independent data cleaning. Despite efforts to clean the data, they exhibit some label noise and may thus benefit from innovative labelling procedures, a challenge we leave to the community. Furthermore, any endeavour dependent on the delineation of categories faces potential biases. Categories simplify and freeze nuanced narratives and obscure political and moral reasoning \cite{Birhane2021}. Despite our cultural domain knowledge, niche categories that remain undiscovered or unavailable online with sufficient images will not be represented in our database. We aim to continue to engage with artists and heritage specialists to improve the representativeness of our categories. 

\paragraph{Ethics and societal impact}

Turath was primarily motivated by the need to increase the cultural diversity of image databases, to improve the applicability of neural networks to under-represented regions, and to actively engage researchers in such regions in the field of machine learning. However, the cultural focus of this database may be prone to abuse by, for example, government and private entities looking to delineate and target cultures for nefarious reasons. To mitigate the abuse of our database for commercial purposes, we are releasing it under a CC BY-NC license, allowing researchers to share and adapt the database in non-commercial settings. More broadly, our belief is that by improving the awareness and understanding of cultures from around the globe, we can better appreciate what they have to offer. Moving forward, we envision the Turath initiative expanding in scope to encompass modalities such as text, audio, and video. Such a path can contribute to research on language preservation, speech recognition, and video analysis.

\bibliographystyle{unsrt}
\bibliography{neurips_2021}

%%%%%%%%%%%%%%%%%%%%%%%%%%%%%%%%%%%%%%%%%%%%%%%%%%%%%%%%%%%%
\section*{Checklist}

\begin{enumerate}

\item For all authors...
\begin{enumerate}
  \item Do the main claims made in the abstract and introduction accurately reflect the paper's contributions and scope?
    \answerYes{We claim and indeed introduce a database (see Sec.~\ref{section:construction}) and evaluate several networks on such a database (see Sec.~\ref{section:results}).}
  \item Did you describe the limitations of your work?
    \answerYes{We discuss the limitations of category definitions and dataset bias (see Sec.\ref{section:discussion})}
  \item Did you discuss any potential negative societal impacts of your work?
    \answerYes{We discuss potential abuse of the dataset by government and non-government entities (see Sec.~\ref{section:discussion})}
  \item Have you read the ethics review guidelines and ensured that your paper conforms to them?
    \answerYes{}
\end{enumerate}

\item If you are including theoretical results...
\begin{enumerate}
  \item Did you state the full set of assumptions of all theoretical results?
    \answerNA{}
	\item Did you include complete proofs of all theoretical results?
    \answerNA{}
\end{enumerate}

\item If you ran experiments...
\begin{enumerate}
  \item Did you include the code, data, and instructions needed to reproduce the main experimental results (either in the supplemental material or as a URL)?
    \answerYes{We include the URL to the corresponding website (which contains code and data) in the abstract. We also include links to demos in Sec.~\ref{section:results}}
  \item Did you specify all the training details (e.g., data splits, hyperparameters, how they were chosen)?
    \answerYes{We include data splits in Table~\ref{table:data_splits}. Implementation details are included in Appendix~\ref{appendix:implementation_details}.}
	\item Did you report error bars (e.g., with respect to the random seed after running experiments multiple times)?
    \answerYes{We report the standard deviation (across five random seeds) of Top-1 and Top-5 accuracy scores in Table~\ref{table:results}.}
	\item Did you include the total amount of compute and the type of resources used (e.g., type of GPUs, internal cluster, or cloud provider)?
    \answerYes{We used Google Colab's GPU resources and outline the duration of each training epoch in Appendix~\ref{appendix:implementation_details}.}
\end{enumerate}

\item If you are using existing assets (e.g., code, data, models) or curating/releasing new assets...
\begin{enumerate}
  \item If your work uses existing assets, did you cite the creators?
    \answerYes{We reference the creators of TensorFlow in Appendix~\ref{appendix:implementation_details}.}
  \item Did you mention the license of the assets?
    \answerYes{We are releasing the database and the code under a CC BY-NC license (see Sec.~\ref{section:discussion})}
  \item Did you include any new assets either in the supplemental material or as a URL?
    \answerYes{We include a link in the abstract to our website which has code, data, and models.}
  \item Did you discuss whether and how consent was obtained from people whose data you're using/curating?
    \answerNA{}
  \item Did you discuss whether the data you are using/curating contains personally identifiable information or offensive content?
    \answerNA{} 
    % We discuss this in Sec.~\ref{section:construction}
\end{enumerate}

\item If you used crowdsourcing or conducted research with human subjects...
\begin{enumerate}
  \item Did you include the full text of instructions given to participants and screenshots, if applicable?
    \answerNA{We did not crowd-source image annotations.}
  \item Did you describe any potential participant risks, with links to Institutional Review Board (IRB) approvals, if applicable?
    \answerNA{Since we did not crowd-source image annotations nor did we involve human subjects, IRB approval was not required.}
  \item Did you include the estimated hourly wage paid to participants and the total amount spent on participant compensation?
    \answerNA{Since we did not involve human participants, payment details are not applicable.}
\end{enumerate}

\end{enumerate}

%%%%%%%%%%%%%%%%%%%%%%%%%%%%%%%%%%%%%%%%%%%%%%%%%%%%%%%%%%%%

\clearpage

\appendix

\section{Database categories}
\label{appendix:database_categories}

In the main manuscript, we described, at a high-level, the contents of the various benchmark databases (Turath Standard, Art, and UNESCO), and outlined the number of image categories that each contains. In this section, we list all the image categories that appear in each of the benchmark databases. Please keep in mind that many of the category names are romanized versions of the original Arabic text, and thus may not be fully comprehensible to non-Arabic speakers.

\subsection{Turath Standard (micro)}

aish el-saraya,
ahaggar national park,
ain ghazal,
ajwa dates,
al-quwaysimah-jordan,
aleppo souk,
aleppo-syria,
alexandria coastline,
alexandria-egypt,
algiers-algeria,
amman-jordan,
ancient jerusalem market,
arabic mamoul food,
ariana-governorate-tunisia,
ayyala folk dance,
babaghanoush,
bamia,
barhi dates,
batna,-algeria-algeria,
beirut-lebanon,
besarah,
bint al sahn,
cairo-egypt,
camel riding,
casablanca-morocco,
cave church egypt,
chorba,
couscous,
damascus-syria,
daraa-syria,
dead sea, jordan,
deir-ez-zor-syria,
desert horse riding dubai,
djelfa-algeria,
dune bashing,
eggah,
egypt basbousa food,
egypt’s black desert,
el mate,
eliyahu hanavi synagogue,
emirate-of-abu-dhabi-the-united-arab-emirates,
emirate-of-fujairah-the-united-arab-emirates,
emirate-of-sharjah-the-united-arab-emirates,
erbil citadel,
essaouira market,
essaouira, morocco,
falafel,
farasan islands, saudi arabia,
farinata,
fasolada,
fatteh,
fattoush,
fesikh,
feteer-meshaltet,
figuig,
freekeh,
ful-medames,
galayet-bandora,
gebel barkal,
giza-egypt,
gouraya national park, algeria,
grape leaves food,
green-beans,
halloumi-cheese,
hama-syria,
haneeth,
harees,
harira,
hawawshi,
hininy,
hummus,
ichkeul lake and national park, tunisia,
idrisid-dynasty-morocco,
iraqi traditional dress,
irbid-jordan,
jabal qara caves,
jeita grotto, lebanon,
jordanian mansaf food,
jordanian traditional dress,
jounieh-lebanon,
kabab,
kabsa,
kairouan-governorate-tunisia,
kamounia,
karak chai,
kebab,
kemençe instrument,
khoshaf,
kibbeh,
kofta,
layali lubnan,
lebanon hummus food,
luqaimat,
mabroom dates,
markook-shrek,
marrakesh-safi-morocco,
medjool dates,
merguez,
merzouga desert,
mesfouf,
mohammad al-amin mosque,
mohammed-ben-abdallah-morocco,
moroccan couscous food,
moussaka,
msemen,
mt. sinai, egypt,
mulukhiyah,
musandam fjords, oman,
musandam oman,
mutabbal,
méchoui,
nile river, egypt,
oasis du sud marocain biosphere reserve,
old mosque of shali fortress,
olives,
omani traditional dress,
oran-algeria,
palestine keffiyeh,
palestine kunafa food,
palestinian maqluba food,
port-said-egypt,
qamar al deen drink,
qualah iraq mountains,
quzi,
rabbi dates,
red sea coast,
rub’ al khali, arabian peninsula,
russeifa-jordan,
sabu-jaddi rock art sites,
safawi dates,
sahlab drink,
saint hilarion monastery,
sandboarding,
saudi kabsa food,
saudi sambousek food,
sayer dates,
sfax-governorate-tunisia,
shishbarak,
shubra-el-kheima-egypt,
sidon-lebanon,
socotra island, yemen,
souk al hamidiyah,
sousse-governorate-tunisia,
sudan traditional dress,
sukkary dates,
syria kibbeh food,
syria qatayef food,
syrian ice cream food,
tabbouleh,
tanbur instrument,
tanger-tetouan-al-hoceima-morocco,
tarim palace yemen,
the church of the annunciation,
tinghir oasis, morocco,
torta-de-gazpacho,
tripoli,-lebanon-lebanon,
tunis-governorate-tunisia,
tyre,-lebanon-lebanon,
wadi mathendous rock art,
wadi rum, jordan,
wadi wurayah biosphere reserve,
waw an namus, libya,
zahidi dates,
zarqa-jordan,
zil instrument,
acacus mountains,
algeria fashion men,
algeria fashion women,
algiers algeria night,
amman jordan night,
arab zaatar,
arabic coffee,
arabic tea,
archery sport,
atlas cedar biosphere reserves,
awamat sweets,
ayran drink,
baalbek-images,
barazik,
beirut lebanon night,
buzuq-images,
cashew fingers,
chrea national park algeria,
constantine algeria,
cracs-images,
dabke dancing,
damascus syria night,
dana biosphere reserve,
derbeke-images,
desert palm tree,
djurdjura national park,
egypt dancing,
egyptian folk dance,
falcon hunting arab gulf,
fez morocco night,
ghraybeh,
giza egypt night,
grand mosque qatar,
hama syria night,
hisham-s palace,
jabal al rihane biosphere reserve,
jabal moussa biosphere reserve,
jarash jordan,
jellab drink,
jet skiing dubai beach,
karkadeh drink,
khan khalil egypt,
khartoum night,
kleicha dessert,
kol w shkor,
kumma hats,
lebanon old houses,
libya fashion women,
madain-images,
marakkesh souq,
marrakech morocco night,
mauritania fashion men,
mauritania fashion women,
mauritania fishing,
mbesses,
meroe-images,
mizmar,
morrocan traditional dress,
muscat capital,
muscat oman night,
muttrah souk,
nay-images,
old souk jeddah,
oman fashion men,
oman fashion women,
omani halwa,
oud-images,
palmyra-images,
petra-images,
qanoon-images,
rabat capital,
ras muhammad national park,
rawshe-images,
rebab,
red sea diving,
riyadh capital,
sanaa yemen night,
santur instrument,
saudi champagne,
saudi male sandals,
saudi old houses,
saudi shemagh,
shamadan dance,
shangeet-images,
sheikh zayed mosque,
shouf biosphere reserve,
subhah beads,
sudan capital,
syria old houses,
table-images,
tamina dessert,
testour mosque,
timgad-images,
traditional fez hat,
tripoli lebanon night,
tunisian dancing,
ula-images,
umm ali dessert,
ummayad mosque,
ummayad-images,
volubilis-images,
yemen fashion men,
yemen fashion women,
yemeni old houses,

\subsection{Turath Art}

abdalla-omari-art,
abdallah-akar-art,
abdallah-benanteur-art,
abdallah-murad-art,
abdel-hadi-el-gazzar-art,
abdel-kader-guermaz-art,
abdel-qader-hassan-art,
abdelkader-benchamma-art,
abdelkebir-rabi-art,
abderrahim-iqbi-art,
abdul-hay-mosallam-zarara-art,
abdul-qader-al-rais-art,
abdul-qadir-al-obaidi-art,
abdul-qadir-al-rassam-art,
abdul-raheem-salem-art,
abdul-rahim-sharif-art,
abdul-rahman-al-maaini-art,
abdul-rahman-mowakket-art,
abdul-rida-bager-art,
abdulhalim-radwi-art,
abdullah-al-muharraqi-art,
abdullah-al-qassar-art,
abdulnasser-gharem-art,
achraf-touloub-art,
adam-henein-art,
adel-abdessemed-art,
adel-abidin-art,
adel-al-khalaf-art,
adel-dauood-art,
adel-el-siwi-art,
adham-wanly-art,
adonis-ali-ahmed-said-esber-art,
afaf-zurayk-art,
afifa-alelby-art,
ahmad-durak-sibai-art,
ahmad-moualla-art,
ahmad-nawash-art,
ahmad-shibrain-art,
ahmed-alsoudani-art,
ahmed-askalany-art,
ahmed-baqer-art,
ahmed-ben-driss-el-yacoubi-art,
ahmed-cherkaoui-art,
ahmed-kassem-art,
ahmed-mater-art,
ahmed-morsi-art,
ahmed-moustafa-art,
ahmed-neshaat-al-zuaby-art,
akram-halabi-art,
akram-zaatari-art,
ala-younis-art,
ali-al-abdan-art,
ali-al-jabri-art,
ali-al-tajer-art,
ali-cherri-art,
ali-ferzat-art,
ali-hassan-art,
ali-mokawas-syria-art,
ali-omar-ermes-art,
ali-rafei-art,
ali-talib-art,
amar-dawood-art,
amer-al-obaidi-art,
ammar-abd-rabbo-art,
ammar-abo-bakr-art,
ammar-al-attar-art,
amr-nazeer-art,
andre-elbaz-art,
armen-agop-art,
asaad-arabi-art,
asim-abu-shakra-art,
asma-fayoumi-art,
atef-maatallah-art,
athar-jaber-art,
atta-sabri-art,
aula-al-ayoubi-art,
aya-tarek-art,
ayad-al-nimar-art,
ayad-alkadhi-art,
ayoub-hussein-art,
baghdad-benas-art,
basel-uraiqat-art,
bashar-alhroub-art,
basim-magdy-art,
bassel-safadi-art,
bassem-dahdouh-art,
batoul-shimi-art,
bibi-zogbe-art,
boushra-al-mutawakel-art,
camille-zakharia-art,
chafic-abboud-art,
chant-avedissian-art,
chaouki-choukini-art,
charbel-joseph-h-boutros-art,
clea-badaro-art,
dana-al-jouder-art,
deirrieh-fakhoury-art,
dia-azzawi-art,
diana-al-hadid-syria-art,
djamel-tatah-art,
djamila-bent-mohamed-art,
driss-ouadahi-art,
ebtisam-abdulaziz-art,
effat-naghi-art,
el-seed-art,
elias-zayat-art,
emmanuel-guiragossian-art,
emmanuel-nassar-art,
ervand-demerdjian-art,
essa-grayeb-art,
etel-adnan-art,
ezequiel-baroukh-art,
fadi-al-hamwi-art,
fadia-haddad-art,
fahr-el-nissa-zeid-art,
faik-hassan-art,
faisal-laibi-sahi-art,
farah-al-qasimi-art,
farah-behbehani-art,
faraj-abbo-al-numan-art,
fares-cachoux-art,
farid-belkahia-art,
farida-el-gazzar-art,
fateh-al-moudarres-art,
fatema-al-mazrouie-art,
fathi-afifi-art,
fathi-hassan-art,
faycal-baghriche-art,
fouad-bellamine-art,
fouad-elkoury-art,
gazbia-sirry-art,
gcc-collective-art,
george-bahgory-art,
george-hanna-sabbagh-art,
ghada-amer-art,
ghadeer-saeed-art,
ghassan-ghaib-art,
ghassan-kanafani-art,
gouider-triki-art,
habib-srour-art,
hadjithomas-joreige-art,
hafidh-aldroubi-art,
haidar-al-mehrabi-art,
halim-al-karim-art,
halim-karibebine-art,
hamdan-al-shamsi-art,
hamed-abdalla-art,
hamed-ewais-art,
hamed-nada-art,
hamza-bounoua-art,
hanaa-malallah-art,
hani-alqam-art,
hani-zurob-art,
hanoos-hanoos-art,
hassan-el-glaoui-art,
hassan-massoudy-art,
hassan-meer-art,
hassan-sharif-art,
hatim-elmekki-art,
hayv-kahraman-art,
hazem-al-zubi-art,
hazem-harb-art,
hazem-mahdy-art,
hedi-turki-art,
helen-khal-art,
hessa-al-joker-art,
hind-nasser-art,
hind-zulfa-art,
huda-lutfi-art,
huguette-caland-art,
hussein-fawzi-art,
hussein-madi-art,
hussein-sharif-art,
hussein-shariffe-art,
ibi-ibrahim-art,
ibrahim-el-salahi-art,
ibrahim-ismail-art,
iman-issa-art,
inaya-fanis-hodeib-art,
inji-efflatoun-art,
ismael-al-khaid-art,
ismail-al-rifai-art,
ismail-fattah-art,
ismail-samson-art,
ismail-shammout-art,
issa-saqer-al-khalaf-art,
issam-al-said-art,
jaber-al-azmeh-art,
jabra-ibrahim-jabra-art,
jafar-islah-art,
jaffar-al-oraibi-art,
jamil-hamoudi-art,
jananne-al-ani-art,
jassim-zaini-art,
jawad-al-malhi-art,
jeffar-khaldi-art,
jewad-selim-art,
jilali-gharbaoui-art,
jorge-tacla-art,
juliana-seraphim-art,
jumana-el-husseini-art,
jumana-manna-art,
kader-attia-art,
kadhim-hayder-art,
kamal-boullata-art,
kamala-ibrahim-ishaq-art,
kamel-el-telmesani-art,
kamel-moghani-art,
kareem-lotfy-art,
kareem-risan-art,
kevork-mourad-art,
khadeir-al-shakarji-art,
khaldoun-shishakly-art,
khaled-al-jader-art,
khaled-hafez-art,
khaled-hourani-art,
khaled-jarrar-art,
khaled-zaki-art,
khalid-al-jallaf-art,
khalid-albaih-art,
khalid-farhan-art,
khalid-mezaina-art,
khalifa-al-qattan-art,
khalil-gibran-art,
khazaal-awad-qaffas-art,
kholoud-al-sharafi-art,
khouzaima-alwani-art,
laila-shawa-art,
lamia-joreige-art,
lamya-gargash-art,
lara-baladi-art,
larissa-sansour-art,
lateefa-bint-maktoum-art,
lawrence-abu-hamdan-art,
layan-shawabkeh-art,
layla-al-attar-art,
layla-juma-art,
leila-nseir-art,
lorna-selim-art,
louay-kayyali-art,
lulwah-al-hamoud-art,
madiha-umar-art,
maha-maamoun-art,
mahmoud-abboud-fahmy-art,
mahmoud-bin-radwan-art,
mahmoud-hammad-art,
mahmoud-obaidi-art,
mahmoud-sabri-art,
mahmoud-said-art,
maitha-demithan-art,
maliheh-afnan-art,
maliheh-afnan-palestine-art,
malika-agueznay-art,
mamdouh-ammar-art,
mamdouh-kashlan-art,
manal-al-dowayan-art,
marguerite-nakhla-art,
mariam-abdel-aleem-art,
marwa-adel-art,
marwa-arsanios-art,
maysa-mohammed-art,
maysaloun-faraj-art,
mazen-ismail-al-ashkar-art,
mejri-thameur-art,
menhat-helmy-art,
michael-rakowitz-art,
michel-basbous-art,
miloud-labeid-art,
moataz-nasr-art,
modhir-ahmed-art,
mohamad-fahmy-ganzeer-art,
mohamad-said-baalbaki-art,
mohamed-abou-el-naga-art,
mohamed-ben-allal-art,
mohamed-chebaa-art,
mohammed-abla-art,
mohammed-ahmed-ibrahim-art,
mohammed-al-kouh-art,
mohammed-al-mazrouie-art,
mohammed-al-qassab-art,
mohammed-farea-art,
mohammed-hamidi-art,
mohammed-ismail-art,
mohammed-issiakhem-art,
mohammed-kacimi-art,
mohammed-kazem-art,
mohammed-khadda-art,
mohammed-mandi-art,
mohammed-masri-art,
mohammed-melehi-art,
mohammed-naghi-art,
mohammed-omar-khalil-art,
mohammed-sabry-art,
mohssin-harraki-art,
mona-hatoum-art,
mona-saudi-art,
moosa-al-halyan-art,
mounirah-mosly-art,
moza-al-suwaidi-art,
muhanna-durra-art,
munira-al-kazi-art,
mustafa-al-hallaj-art,
nabil-nahas-art,
nabil-safwat-art,
nadia-ayari-art,
nadia-kaabi-linke-art,
nadia-saikali-art,
nadim-raef-art,
naim-ismail-art,
najat-maki-art,
najla-al-saleem-art,
nasser-al-yousif-art,
nazar-yahya-art,
naziha-selim-art,
nazir-ismail-art,
nazir-nabaa-art,
nedim-kufi-art,
nejib-belkhoja-art,
nermine-hammam-art,
nidhal-chamekh-art,
nja-mahdaoui-art,
noor-al-suwaidi-art,
noor-bahjat-art,
nouri-al-rawi-art,
obaid-suroor-art,
omar-al-rashid-art,
omar-el-nagdi-art,
omar-hamdi-art,
omar-khairy-art,
omar-onsi-art,
paul-guiragossian-art,
raafat-ishak-art,
rachid-koraichi-art,
rafa-al-nasiri-art,
rafic-charaf-art,
ragheb-ayad-art,
rajiha-qudsi-art,
ramses-younan-art,
rashid-al-oraifi-art,
rawya-ahmed-malik-art,
reda-abdelrahman-art,
reem-al-faisal-art,
reem-al-ghaith-art,
rim-al-jundi-art,
saad-ben-cheffaj-art,
saad-el-khadem-art,
saadi-al-kaabi-art,
sadik-alfraji-art,
safia-farhat-art,
safwan-dahoul-art,
salah-abdel-kerim-art,
salah-taher-art,
salama-safadi-art,
saleh-al-jumaie-art,
saliba-douaihy-art,
salman-abbas-art,
salman-al-basri-art,
saloua-raouda-choucair-art,
sama-al-shaibi-art,
sami-mohammed-art,
samia-halaby-art,
samir-rafi-art,
samir-sayegh-art,
samira-badran-art,
seif-wanly-art,
seta-manoukian-art,
shaaban-zaki-art,
shada-safadi-art,
shadi-alzaqzouq-art,
shadi-habib-allah-art,
shakir-hassan-al-said-art,
sharif-waked-art,
shawki-youssef-art,
simone-fattal-art,
sinan-hussein-art,
sophia-al-maria-art,
steve-sabella-art,
suad-al-attar-art,
sueraya-shaheen-art,
suha-shoman-art,
sulafa-hijazi-art,
suleiman-mansour-art,
susan-hefuna-art,
tagreed-darghouth-art,
tahia-halim-art,
talal-moualla-art,
tammam-al-akhal-art,
tammam-azzam-art,
tarek-al-ghoussein-art,
tawfik-al-alousi-art,
tayseer-barakat-art,
taysir-batniji-art,
thuraya-al-baqsami-art,
ufemia-rizk-art,
van-leo-art,
vera-tamari-art,
wael-darwish-art,
walead-beshty-art,
walid-al-shami-art,
walid-ebeid-art,
walid-raad-art,
walid-siti-art,
waseem-marzouki-art,
wassef-boutros-ghali-art,
wijdan-ali-art,
yasser-dweik-art,
yasser-rostom-art,
yousef-ahmed-art,
youssef-kamel-art,
youssef-nabil-art,
yto-barrada-art,
yvette-achkar-art,
zena-al-khalil-art,
zena-assi-art,
zhivago-duncan-art,
ziad-antar-art,
ziad-dalloul-art,
zineb-sedira-art,
zoulikha-bouabdellah-art,

\subsection{Turath UNESCO}

abu-mena-unesco-site,
aflaj-irrigation-systems-of-oman-unesco-site,
ahwar-of-southern-iraq-unesco-site,
al-ahsa-oasis-unesco-site,
al-ain-unesco-site,
al-balad,-jeddah-unesco-site,
al-maghtas-unesco-site,
al-zubarah-unesco-site,
amphitheatre-of-el-jem-unesco-site,
ancient-city-of-bosra-unesco-site,
ancient-city-of-damascus-unesco-site,
ancient-ksour-of-ouadane,-chinguetti,-tichitt-and-oualata-unesco-site,
anjar,-lebanon-unesco-site,
archaeological-site-of-carthage-unesco-site,
archaeological-sites-of-bat,-al-khutm-and-al-ayn-unesco-site,
assur-unesco-site,
baalbek-unesco-site,
babylon-unesco-site,
bahla-fort-unesco-site,
bahrain-pearling-trail-unesco-site,
battir-unesco-site,
beni-hammad-fort-unesco-site,
byblos-unesco-site,
casbah-of-algiers-unesco-site,
cedars-of-god-unesco-site,
church-of-the-nativity-unesco-site,
citadel-of-arbil-unesco-site,
citadel-of-salah-ed-din-unesco-site,
cyrene,-libya-unesco-site,
dead-cities-unesco-site,
dilmun-burial-mounds-unesco-site,
diriyah-unesco-site,
djémila-unesco-site,
dougga-unesco-site,
el-jadida-unesco-site,
essaouira-unesco-site,
fes-el-bali-unesco-site,
frankincense-trail-unesco-site,
gebel-barkal-and-the-sites-of-the-napatan-region-unesco-site,
ghadames-unesco-site,
giza-pyramid-complex-unesco-site,
hatra-unesco-site,
hebron-unesco-site,
ichkeul-national-park-unesco-site,
islamic-cairo-unesco-site,
kadisha-valley-unesco-site,
kairouan-unesco-site,
kerkouane-unesco-site,
krak-des-chevaliers-unesco-site,
ksar-of-ait-ben-haddou-unesco-site,
leptis-magna-unesco-site,
medina-of-marrakesh-unesco-site,
medina-of-sousse-unesco-site,
medina-of-tunis-unesco-site,
meknes-unesco-site,
meroë-unesco-site,
necropolis-of-kerkouane-unesco-site,
nubian-monuments-from-abu-simbel-to-philae-unesco-site,
old-city-of-aleppo-unesco-site,
petra-unesco-site,
qalhat-unesco-site,
qasr-amra-unesco-site,
rabat-unesco-site,
rock-art-sites-of-tadrart-acacus-unesco-site,
sabratha-unesco-site,
samarra-unesco-site,
shibam-unesco-site,
site-of-palmyra-unesco-site,
theban-necropolis-unesco-site,
thebes,-egypt-unesco-site,
timgad-unesco-site,
tipaza-unesco-site,
tyre,-lebanon-unesco-site,
tétouan-unesco-site,
umm-ar-rasas-unesco-site,
volubilis-unesco-site,
wadi-al-hitan-unesco-site,
wadi-rum-unesco-site,
zabīd-unesco-site,

\section{Implementation details}
\label{appendix:implementation_details}

To allow for the reproducibility of our image classification experiments, we outline, in Table~\ref{table:implementation_details}, the implementation details of those experiments. We use TensorFlow \cite{Abadi2016} for all experiments and during hyperparameter optimization, we experimented with learning rates in the range $lr \in [1e^{-4} - 1e^{-3}]$. We did not implement any data augmentation strategy during training, such as random cropping, rotations, etc. All images were reshaped to $224 \times 224$ before being fed to a network. For all experiments, and to mitigate over-fitting, we implemented an early stopping criterion based on the loss incurred on the validation set with a patience value of 5 epochs. For evaluation purposes, we extracted and exploited the parameters that coincided with the minimum loss incurred on the validation set. The experiments leveraged the GPU resources on Google Colab and, depending on the benchmark database, each epoch of training and evaluation on the validation set was $30-200s$ in duration. 

\begin{table}[!h]
    \centering
    \caption{\textbf{Implementation details of the image classification experiments conducted on the benchmark databases.} LR and BS refer to the learning rate and batch-size respectively. Macro and micro refer to the granularity of the category labels used during training and evaluation.}
    \label{table:implementation_details}
    \begin{tabular}{c|c c c c}
         \toprule
         Benchmark & Optimizer & Loss & LR & BS \\
         \midrule
         Turath Standard \textit{(macro)} & Adam & Cross-entropy & $1e^{-3}$ & 64 \\
         Turath Standard \textit{(micro)} & Adam & Cross-entropy & $1e^{-4}$ & 64 \\
         Turath Art & Adam & Cross-entropy & $1e^{-4}$ & 64 \\
         Turath UNESCO & Adam & Cross-entropy & $1e^{-4}$ & 64 \\
         \bottomrule
    \end{tabular}
\end{table}

\section{Limitations of networks pre-trained on ImageNet}
\label{appendix:network_limitations}

\begin{figure}[!h]
    \centering
    \begin{subfigure}{\textwidth}
        \includegraphics[width=\textwidth]{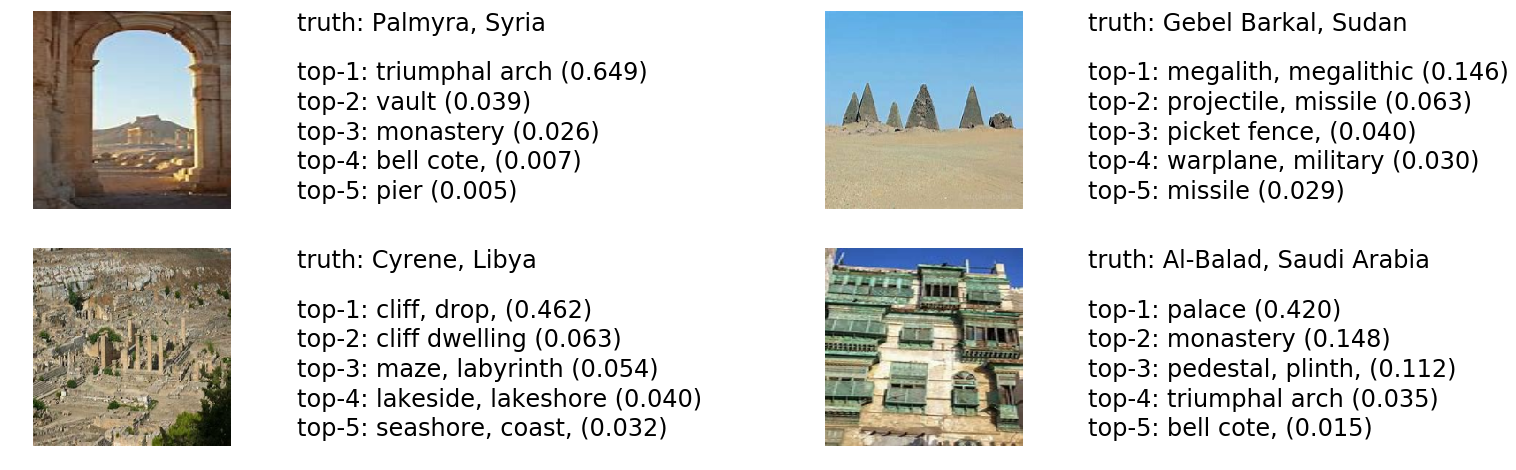}
        \caption{MobileNetV2}
        \label{fig:failures_standard_mobilenet}
    \end{subfigure}
    ~
    \begin{subfigure}{\textwidth}
        \includegraphics[width=\textwidth]{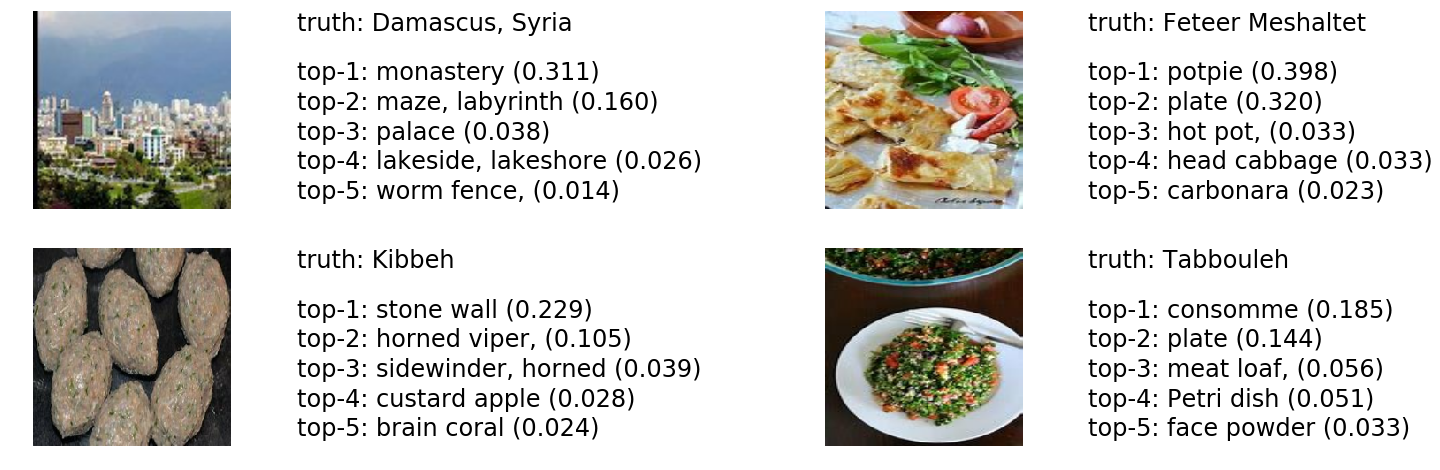}
        \caption{ResNet50}
        \label{fig:failures_standard_resnet}
    \end{subfigure}
    \caption{\textbf{Top-5 predictions (and confidence) made by networks pre-trained on ImageNet and directly deployed on image samples from the Turath Standard benchmark.} We also present the ground-truth \textit{micro} category of each of the image samples. Most of the predictions are incorrect, lack the finer resolution of our micro categories, and do not have a cultural emphasis.}
    \label{fig:failures_standard}
\end{figure}

In the main manuscript, we made the case for the limitations of networks pre-trained on ImageNet. We did so by deploying an EfficientNet on image samples from the Turath database and comparing the Top-5 predictions to the ground-truth label. In this section, we extend those findings to other neural architectures, including MobileNetV2 and ResNet50. We randomly sample 9 images from the Turath database, perform a forward pass through the network, and present the Top-5 predictions and corresponding confidence levels in Figs~\ref{fig:failures_standard_mobilenet} and \ref{fig:failures_standard_resnet}.  

We find that, regardless of the neural architecture, networks pre-trained on ImageNet are unable to correctly predict the micro-level category of image samples from the Turath database. For example, in Fig.~\ref{fig:failures_standard_mobilenet}, we see that MobileNetV2 misclassifies $\mathrm{Cyrene}$, an ancient Greek city in present-day Libya, as a $\mathrm{cliff}$. Similarly, it misclassifies $\mathrm{Gebel \ Barkal}$, pyramids in present-day Sudan, as a $\mathrm{megalith}$. In Fig.~\ref{fig:failures_standard_resnet}, we see that ResNet50 confidently misclassifies a scene from $\mathrm{Damascus, \ Syria}$, as a $\mathrm{monastery}$ and confuses $\mathrm{Kibbeh}$, a traditional Arab food item, for a $\mathrm{stone \ wall}$. 

\end{document}